\documentclass[10pt,twocolumn,twoside]{IEEEtran}
\usepackage{amsmath,amsfonts}
\usepackage{amsthm,amsmath,amssymb}
\usepackage{mathrsfs}
\usepackage{array}
\usepackage{textcomp}
\usepackage{stfloats}
\usepackage{url}
\usepackage{verbatim}
\usepackage{graphicx}
\graphicspath{ {./images/} }
\usepackage{epstopdf}

\usepackage[table]{xcolor}
\def\BibTeX{{\rm B\kern-.05em{\sc i\kern-.025em b}\kern-.08em
    T\kern-.1667em\lower.7ex\hbox{E}\kern-.125emX}}
\usepackage{balance}
\usepackage[labelformat=simple]{subcaption}

\usepackage{multirow}
\usepackage{booktabs}
\usepackage{xcolor}
\usepackage{braket}
\usepackage{hyperref}
\usepackage{tikz}
\usepackage{graphicx}
\usepackage{algorithm}
\usepackage[noend]{algpseudocode}
\makeatletter
\renewcommand{\maketag@@@}[1]{\hbox{\m@th\normalsize\normalfont#1}}%
\makeatother

\begin{document}

\title{Distribution-Level Memory Recall for Continual Learning: Preserving Knowledge and Avoiding Confusion}
\author{
Shaoxu Cheng, Kanglei Geng, Chiyuan He, Zihuan Qiu, Linfeng Xu$ ^{\ast}$, Heqian Qiu, Lanxiao Wang, Qingbo Wu, Fanman Meng, Hongliang Li
 \thanks{The authors are with the School of Information and Communication Engineering, University of Electronic Science and Technology of China, Chengdu 611731, China (e-mail: shaoxu.cheng@std.uestc.edu.cn; kangleigeng@163.com; cyhe@std.uestc.edu.cn; zihuanqiu@std.uestc.edu.cn; lfxu@uestc.edu.cn; hqqiu@uestc.edu.cn; lanxiao.wang@std.uestc.edu.cn; qbwu@uestc.edu.cn; fmmeng@uestc.edu.cn; hlli@uestc.edu.cn}
}

\maketitle

\begin{abstract}
Continual Learning (CL) aims to enable Deep Neural Networks (DNNs) to learn new data without forgetting previously learned knowledge. The key to achieving this goal is to avoid confusion at the feature level, i.e., avoiding confusion within old tasks and between new and old tasks. Previous prototype-based CL methods generate pseudo features for old knowledge replay by adding Gaussian noise to the centroids of old classes. However, the distribution in the feature space exhibits anisotropy during the incremental process, which prevents the pseudo features from faithfully reproducing the distribution of old knowledge in the feature space, leading to confusion in classification boundaries within old tasks. To address this issue, we propose the Distribution-Level Memory Recall (DMR) method, which uses a Gaussian mixture model to precisely fit the feature distribution of old knowledge at the distribution level and generate pseudo features in the next stage. Furthermore, resistance to confusion at the distribution level is also crucial for multimodal learning, as the problem of multimodal imbalance results in significant differences in feature responses between different modalities, exacerbating confusion within old tasks in prototype-based CL methods. Therefore, we mitigate the multi-modal imbalance problem by using the Inter-modal Guidance and Intra-modal Mining (IGIM) method to guide weaker modalities with prior information from dominant modalities and further explore useful information within modalities. For the second key, We propose the Confusion Index to quantitatively describe a model's ability to distinguish between new and old tasks, and we use the Incremental Mixup Feature Enhancement (IMFE) method to enhance pseudo features with new sample features, alleviating classification confusion between new and old knowledge. We conduct extensive experiments on the CIFAR100, ImageNet100, and UESTC-MMEA-CL datasets, and achieve state-of-the-art results.
\end{abstract}


\section{Introduction}
\IEEEPARstart{D}{eep} neural networks (DNNs) have excelled in learning complex patterns and representations from data, leading to numerous breakthroughs in industry \cite{diffusion,vit,tmm2,tmm1,tmm3}. However, complex data emerges continuously in the form of data streams. In this scenario, deep neural networks (DNNs) have been found to struggle with processing continuous data. One of the challenges in training DNNs is the phenomenon where they tend to forget previously learned knowledge when new data is introduced, known as ``catastrophic forgetting''. Various methods have been proposed to address this issue from different perspectives. One approach involves storing a small portion of old samples while continuously inputting new samples into the network. Another method involves using knowledge distillation to constrain the student model to retain old knowledge at the input-output level. Regularizing network parameters, continually expanding and adjusting network structures, and other methods focus on preserving old knowledge by manipulating model parameters or structures. Prototype-based methods also exist, which maintain old knowledge by using prototypes of old categories. The exemplar-free prototype-based method is favored for its privacy-preserving characteristics. However, it lacks a thorough consideration from the perspective of preserving classification boundaries of old tasks. Prior prototype-based methods generate pseudo features in the incremental task phase that cannot accurately reproduce old knowledge, leading to confusion in classification boundaries within old tasks. Therefore, for prototype-based methods, instead of combining them with more complex network structures, regularization strategies, or using networks from both old and new stages for knowledge distillation, it is more effective to seek solutions from the distribution of old knowledge itself. 
We propose a method for preserving and reproducing knowledge at the distribution level in the representation space. At each stage, we use GMM to fit the distribution of high-dimensional features, and improve it by adaptively determining the number of Gaussian components. Furthermore, we apply twice information degradation to preserve the distribution. In the next task stage, we generate pseudo features. The resulting pseudo features closely approximate the true distribution of old samples, maintaining the classification boundaries between classes within the old task and maximizing the preservation of ``true memories''. Furthermore, Artificial intelligence has made significant breakthroughs in the industrial sector, sparking a wave of applications \cite{mm1, mm2, mm3, mm4}. We continuously study its application in real-world scenarios of multimodal behavior recognition. Building on Peng et al.'s work \cite{Peng_2022_CVPR}, we delve into the issue of imbalance in multimodal networks, which can lead to suboptimal performance. We observe that the imbalance between modalities results in significant differences in the response of different modal features \ref{balance}, rendering previous prototype-based methods ineffective for reproducing old knowledge. Therefore, we propose a dual information enhancement approach within and between modalities to address the imbalance caused by the insufficient information in weaker modalities. When our method can maintain the feature distribution and classification boundaries of the old task, and the imbalance in the multimodal distribution can be alleviated, we observe through the Confusion Index that the confusion between the new and old tasks, leading to classification errors, cannot be ignored. Therefore, we use features of the new task's samples to perform data augmentation on pseudo features of the old task's knowledge to alleviate the confusion between the new and old tasks.

In summary, to address the aforementioned challenges, we categorize the paradigm of human intuition overcoming forgetfulness into three aspects, which serve as the overall framework. Firstly, it is essential to retain true old knowledge to ensure that the replayed old knowledge aligns as closely as possible with reality. Secondly, there must be a learning ability, which is also a prerequisite for maintaining valuable memory. In the case of multimodal scenario applications, the imbalance of multimodalities may fail to meet this prerequisite. Finally, there should be no confusion between new and old knowledge. We develop strategies to alleviate the forgetting problem based on this paradigm:
 \begin{figure}[t]
    \centering
    \includegraphics[width=0.9\linewidth]{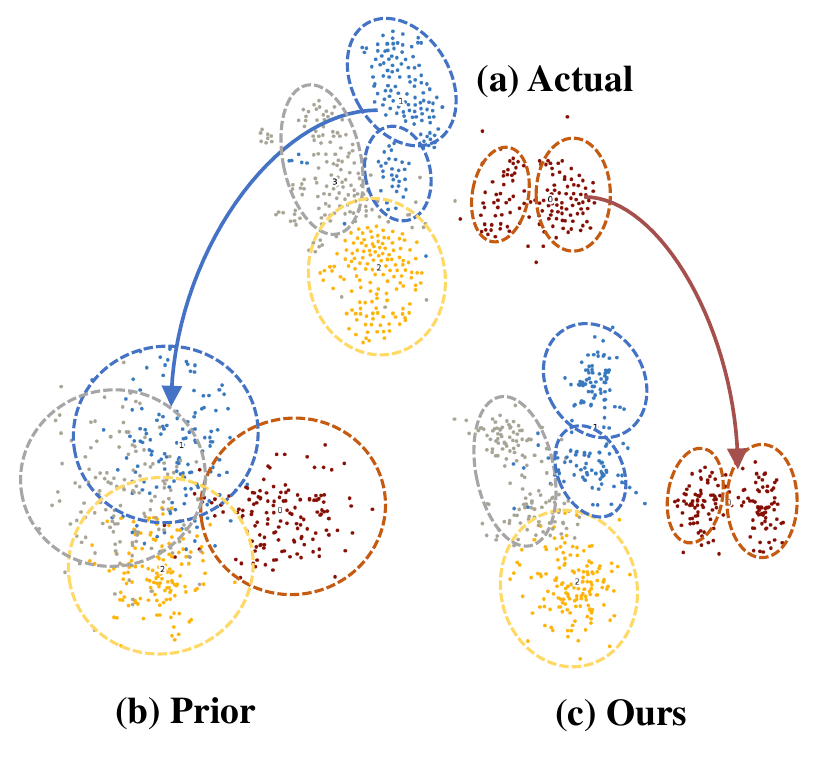}
    \caption{The distribution of features after dimensionality reduction using t-SNE \cite{van2008visualizing}, where (a) represents the actual embedding space distribution of samples from four classes; (b) represents the pseudo features generated during the incremental stage using the class centers and deviations stored in (a); and (c) represents the pseudo features generated during the incremental stage using our method.} 
    \label{tsne_actual_vs_gen} 
    \vspace{-1em}
\end{figure}

\textbf{1) Memories must not ``deform'':} For the incremental-frozen \cite{fantasy} method, the distribution of the feature space exhibits anisotropy \cite{goswami2024fecam}, making it extremely difficult for the original prototype-based method to maintain knowledge from previous tasks solely by relying on simple data from class centers. As shown in Fig. \ref{tsne_actual_vs_gen}, subfigure (a) shows the distribution of features obtained from real samples from four classes in the embedding space, (b) simulates the pseudo features generated in the feature space based on class centers and mean squared deviations alone, and (c) shows the simulated results of (a) using our method. It can be observed that generating data solely based on class centers cannot restore the feature space distribution of the old task, loses the knowledge learned within classes, and confuses the decision boundary. Therefore, we propose the Distribution-level Memory Recall (DMR) method to fit the real feature space distribution, better preserving the knowledge learned within classes in the old task, and better restoring the decision boundaries between classes in the old task.

\textbf{2) Overcoming confusion:} In the context of feature space, the memorization of old classes is already established. Therefore, distinguishing between old and new knowledge is crucial when acquiring new knowledge, as otherwise, there is a risk of confusion between the two. To address this issue, we propose the Incremental Mixup Feature Enhancement (IMFE) method. This method utilizes the features of new classes to enhance pseudo features, thereby improving the classifier's ability to distinguish between old and new knowledge.

\textbf{3) Enhancing learning capabilities:} We propose the Inter-Modal Guidance and Intra-Modal Mining (IGIM) method, which enhances information by considering both inter- and intra-modal aspects. Our approach first categorizes modalities into dominant and weak based on their performance,  we select prior effective knowledge to guide the learning of weak modalities. Additionally, we deeply mine useful information within each modality. These dual approaches reduce the gap between dominant and weak modalities, leading to more generalizable features.

Our contributions can be summarized as follows:
\begin{itemize}
\item  DMR method we proposed is the first to fit the distribution of representations in the feature space from the perspective of maintaining the old knowledge distribution and thereby maintaining the classification boundaries within the old task. We also propose an optimization plan for the fitted storage burden, which can ultimately maintain realistic classification boundaries with almost no increase in burden.
\item  By transferring prior information from dominant modalities and mining the time-frequency information of sensor modalities, we alleviate the imbalance between modalities and obtain generalizable features, meeting the prerequisites of multimodal continual learning.
\item  Observation is conducted through a customized confusion index, and the classifier's ability to distinguish between new and old knowledge is enhanced using the IMFE method, reducing the confusion between new and old classes.
\item We conducted extensive experiments on the CIFAR100 \cite{krizhevsky2009learning(cifar)}, ImageNet100 \cite{russakovsky2015imagenet, Rebuffi17(icarl)}, and UESTC-MMEA-CL \cite{xu2023towards} datasets, achieving SOTA results, and performed detailed ablation studies and analysis.
\end{itemize}

This paper is an extended version of our previous work reported at \cite{csx}, with the following major improvements compared to the previous version: (1) From the perspective of knowledge distribution, we employ GMM to protect knowledge at the distribution level and use an adaptive algorithm to calculate the number of feature subspaces. We also use dual information degradation to avoid storage overhead. We evaluate and analyze the similarity between pseudo-features and the original real distribution. (2) By using the Confusion Index between new and old categories, we analyze the confusion between new and old categories, providing a basis for the methods in the original work. (3) We extend the original work to the theory of continual learning and propose a paradigm for continual learning, demonstrating its effectiveness on more general datasets. (4) In multi-modal applications, we address the issue of multi-modal imbalance by guiding weak modalities with prior knowledge to achieve deeper solutions.

\section{Related Work}
\subsection{Continual Learning}
In the realm of continual learning, the goal is to enable artificial neural networks to continuously ``evolve'' in their learning, acquiring new knowledge without forgetting. To mitigate forgetting, several approaches have been proposed:

First is the Dynamic Network Approaches, they extend the network backbone or some network layers, enabling the extended parts to accommodate new knowledge while the original parts maintain old knowledge. DEN \cite{yoon2017lifelong(DEN)} trains neurons relevant to incremental tasks purposefully, mitigating the drift of individually optimized neurons. DER \cite{Yan21(DER)} also expands the backbone continuously during incremental learning to retain old knowledge, but with a broader linear classifier aggregating features from multiple backbones. Inspired by gradient boosting algorithms, FOSTER \cite{wang2022foster} dynamically adjusts the network and employs knowledge distillation to retain old knowledge. However, these methods may face issues with excessive memory consumption due to the continual expansion of the backbone network. MEMO \cite{zhou2022model(MEMO)} observes that only deeper network layers exhibit specificity to different incremental tasks, thus saving storage by expanding only specified network layers.

Parameter regularization methods impose regularization on network parameters to prevent parameter drift, EWC \cite{Kirkpatrick17(ewc)} evaluates the importance of parameters for old knowledge using the Fisher information matrix, protecting parameters deemed more crucial for old knowledge, thus safeguarding old knowledge from a parameter perspective. SI \cite{Zenke17(SI)} defines the importance of parameters for old knowledge based on the magnitude of their influence on the loss function, proposing an online method for importance assessment. IADM \cite{yang2019adaptive(IADM)} assesses the importance of parameters in different network layers, contrary to EWC \cite{Kirkpatrick17(ewc)}, which regularizes global parameters.

Knowledge distillation is a mainstream method in continual learning. These methods add regularization distillation loss functions to the inputs and outputs. LwF \cite{li2017learning(LWF)} was the first to apply distillation strategies in continual learning. It maintains old knowledge by using the frozen old model as the teacher model to guide the student model in the incremental phase at the input and output levels. The iCaRL \cite{Rebuffi17(icarl)} builds upon LwF \cite{li2017learning(LWF)} by adding an exemplar set for replay to further protect old knowledge. COIL \cite{zhou2021co(COIL)} employs bidirectional distillation to leverage semantic relationships in both new and old models.

In addition to logits distillation, UCIR \cite{hou2019learning(UCIR)} employs a more potent feature distillation, stipulating that the encoder's outputs should be similar. It discovers weight drift in the classifier and addresses this issue by using a cosine classifier. Building upon this, WA \cite{zhao2020maintaining(WA)} continuously normalizes weights and introduces weight clipping to prevent drift. ISM-Net \cite{qiu2023ism} combines distillation with incremental semantic mining between taskes, thereby continuously retaining old knowledge while expanding the network to learn new knowledge.
PODNet \cite{Douillard20(PODNet)} pools and reduces the differences in feature maps from different network layers before distillation. IL2A \cite{zhu2021class(IL2A)} evaluates the feature values obtained from the spectral decomposition of features of new and old classes, finding that maintaining relatively large feature values can mitigate forgetting, further alleviated by distillation.

These methods can be classified into two classes: replay-based and exemplar-free methods. Strategies involving replay exemplars are often combined with other approaches, such as the FOSTER \cite{wang2022foster} combining with dynamic networks and the LwM \cite{dhar2019learning(LwM)} method combining with parameter regularization, especially in conjunction with distillation. Because these methods can directly access old data during the incremental process, they generally outperform methods that do not use replay strategies.

Prototype-based methods primarily operate in the embedding space, without direct access to the raw data of old classes. However, when IL2M \cite{belouadah2019il2m(IL2M)} first introduced methods that store prototypes and simultaneously use sample replay strategies, subsequent prototype-based methods mostly avoided using replay strategies. The PASS \cite{zhu2021prototype(pass)} first obtains a generalized encoder and features through self-supervised methods in the initial stage. It then stores the class centers and mean squared deviations of old classes, and enhances the prototypes with Gaussian noise in the incremental stage to preserve old knowledge. Subsequent SSRE \cite{zhu2022self(SSRE)} method selects samples based on the similarity between prototypes and new samples, and then distills to enhance the Discriminability between new and old classes. Additionally, it extends and reorganizes network layers to maintain old features. The FeTrIL \cite{petit2023fetril} enhances old prototypes by computing the similarity between new and old class prototypes and shifting all features of new samples to the positions of old class prototypes, thereby maintaining knowledge of old classes.
Prototype-based methods can maintain knowledge through prototypes of old classes at the feature level.

\subsection{Multimodal Continual Learning}
In addition to the problem of class imbalance, the work on multimodal continual learning has also attracted considerable attention, especially in the field of vision and language. Srinivasan et al. \cite{srinivasan2022climb(CILMB)} proposed the CLiMB framework, which modifies the vision-language transformer to implement several continual learning methods. They provide a benchmark for continual learning in vision-language tasks, aiming to promote research on novel continual learning algorithms in this challenging multimodal environment. Zhu et al. \cite{zhu2023ctp(CTP)} contributed a unified multimodal dataset for continual learning. They also introduced the Compatible Momentum Contrast method for Topology Preservation, which updates the features of modalities by absorbing knowledge from old and new tasks separately. 

The popular vision-language pretraining model CLIP has also garnered attention. Wang et al. \cite{wang2022continual(Continual)} proposed a continual learning model based on the CLIP model, which uses probabilities for fine-tuning to alleviate forgetting issues. It demonstrates outstanding performance in detecting new data and example selection. The S-prompt \cite{wang-s-prompt} method explores domain-incremental learning using the CLIP model but does not address class incremental learning.

With the development of sensors and other devices, in the field of multimodal continual activity recognition, HarMl \cite{zhang2021harmi(HarMI)} considered the significant storage consumption in training multimodal models and proposed corresponding optimizations. They used an attention mechanism to handle data from different sensors and also employed elastic weight consolidation and exemplar relevance analysis to overcome catastrophic forgetting. Xu et al. \cite{xu2023towards} found that multimodal data exacerbates the problem of forgetting in continual learning, and different modal fusion methods have different effects on the forgetting problem. Modal fusion makes features contain more useful information, enhancing the expressive power of each modality. He et al. \cite{he2024continual} found that by masking the representation of dominant modalities in multimodal data and prolonging the optimization time of weaker modalities, the forgetting problem in the incremental process can be effectively mitigated. Wang et al. \cite{wanghanxin} generated confusion samples through mixup \cite{zhang2017mixup} to encourage the fusion network to learn more discriminative features to alleviate feature confusion between different tasks. AID \cite{csx} used a time-frequency attention mechanism for sensor signals, thus alleviating unbalance, and used a mix of new and old classes of features and pseudo features to reduce confusion between old and new tasks.
In the AV-CIL \cite{pian2023audio(AV-CIL)}, benchmarks were established on three multimodal datasets using pretrained models. They maintained semantic similarity between modalities by constraining the similarity between two modalities and used distillation methods to alleviate forgetting problems.

\section{Methodology}

\subsection{Problem Statement}
Continual learning is a learning paradigm that requires data to be continuously inputted as a series of tasks, mimicking the scenario of continually receiving new data for learning. We partition the data into different tasks $\mathcal{D} = \left\{ \mathcal{D}_{\tau}\right\}_{\tau=1}^{T}$, where $\mathcal{D_\tau}=\left\{ x_{\tau,i},y_{\tau,i}\right\}_{i=1}^{N_{\tau}}$ represents the input of $N_{\tau}$ data $x_{\tau,i} \in \mathbb{R}^{D}$ at stage $\tau$, with the corresponding label $y_{\tau,i} \in C_{\tau}$, where $C_{\tau}$ is the class set in task $\tau$, and different stages satisfy $C_{t_{1}}\cap C_{t_{2}}=\emptyset$. We decouple the entire model into an encoder and a classifier:
$f\left( x_{\tau,i} \right)=W^{\top} \phi(x_{\tau,i}),$
where $\phi(\cdot):\mathbb{R}^{D} \to \mathbb{R}^d, W \in \mathbb{R}^{d \times  |y_{\tau}|}$.

\subsection{Distribution-level Memory Recall}
\label{DMR}
\begin{figure}[h]
    \centering
    \includegraphics[width=\linewidth]{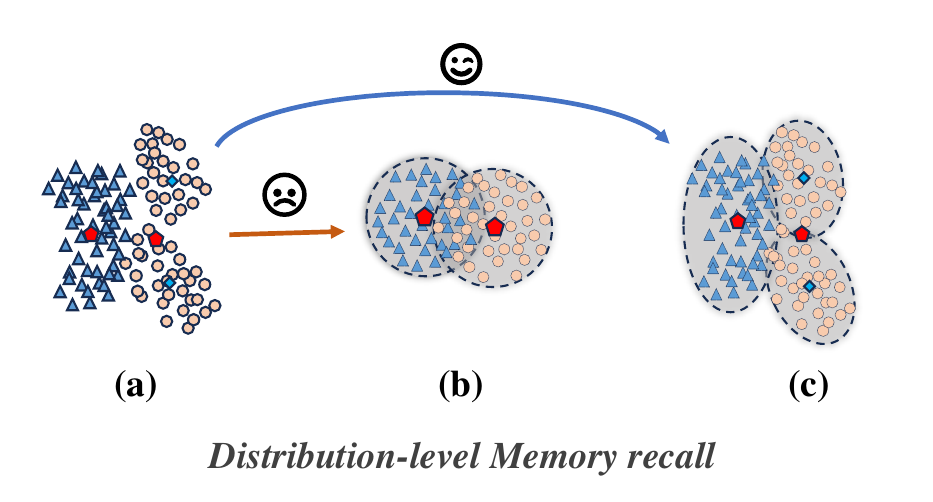}
    \caption{Illustration of the motivation behind the distribution-level memory recall method. We aim to avoid significant information loss and inter-class confusion when old knowledge is reproduced. Instead, we prefer to maintain its original distribution when recalling memories.} 
    \vspace{-1em}
    \label{memory-recall} 
\end{figure}
\textbf{Revisiting Prototype-Based CIL Methods: Transitioning from (a) to (b),}
Prototype-based methods, which do not involve privacy concerns and can preserve old knowledge similar to rehearsal-based methods, are widely used in continual learning and few-shot continual learning. For example, PASS \cite{zhu2021prototype(pass)} obtains pseudo features in the class-incremental phase by using class centers plus the mean squared deviation multiplied by Gaussian noise. In FeTrIL \cite{petit2023fetril}, all features of new classes are shifted to the class centers of similar old classes to obtain pseudo features of old samples. However, these methods fail to consider that the samples of new classes exhibit an anisotropic distribution due to fixed encoder parameters, and there are inherent differences in distributions between new and old classes. This can lead to confusion in the classification boundary when shifting, resulting in incomplete preservation of old knowledge.
Fig. \ref{memory-recall} is a schematic diagram related to Fig. \ref{tsne_actual_vs_gen}, where (a) represents the actual distribution of feature space for two classes, and (b) illustrates the method of enhancing class centers with Gaussian noise to generate pseudo features. It can be seen that this method fails to effectively reproduce the actual situation in (a), leading to confusion in the classification boundary between classes, indicating incomplete preservation of knowledge from the old classes.

\textbf{How Does Our Method Progress from (a) to (c)?}
To address the above issues, we first aim to preserve complete old knowledge, as shown in subfigure (c), to maintain more accurate old knowledge and thus maintain approximately true inter-class relationships when generating pseudo features in the incremental phase. We propose the Distribution-level Memory Recall method to effectively preserve the feature space distribution of old knowledge and prevent inter-class confusion.

We fit all features in the incremental phase of task $\tau$ using GMM. Assuming that the feature distribution in incremental learning can be modeled and fitted by a mixture model consisting of $K$ Gaussian distributions, and there is a dataset containing n samples $\{x_1, x_2, ..., x_n\}$, where each sample $x_i$ is a d-dimensional vector (i.e., $x_i=(x_{i1},x_{i2},...,x_{id})$).
Let's assume that the probability density function (PDF) of the Gaussian distribution corresponding to the $k$-th ($0<k \le K$) mixture component is $N(\mu_k, \Sigma_k)$, where $\mu_k$ is a d-dimensional mean vector, and $\Sigma_k$ is a $d\times d$ covariance matrix. Then, the PDF of the $k$-th mixture component can be represented as:
\begin{scriptsize}
    \begin{equation}
        p(x|\mu_k, \Sigma_k) = \frac{1}{(2\pi)^{d/2}|\Sigma_k|^{1/2}} \exp\left(-\frac{1}{2}(x-\mu_k)^T\Sigma_k^{-1}(x-\mu_k)\right),
    \end{equation}
\end{scriptsize}  
the PDF of the entire GMM can be expressed as a linear combination of the PDFs of each mixture component:
\begin{equation}
    p(x) = \sum_{k=1}^{K} \pi_k p(x|\mu_k, \Sigma_k),
\end{equation}
where $\pi_k$ is the weight of the k-th mixture component, satisfying $\sum_{k=1}^{K} \pi_k = 1$.
Our goal is to estimate the parameters of the GMM by maximizing the joint PDF of all samples in the dataset, i.e., maximizing the likelihood function:
\begin{equation}
    L(\Theta) = \prod_{i=1}^{n} p(x_i|\Theta),
\end{equation}
where $\Theta$ represents all parameters of the GMM, including the means $\mu_k$, covariance matrices $\Sigma_k$, and weights $\pi_k$ of each mixture component.

In practice, to avoid numerical issues, the logarithm of the likelihood function is often taken, and the negative is minimized, transforming the problem of maximizing the likelihood function into minimizing the negative log-likelihood function $\text{argmin}_\Theta (-\log L(\Theta))$:
    \begin{equation}
       \text{argmin}_\Theta \left(-\sum_{i=1}^{n} \log \left(\sum_{k=1}^{K} \pi_k p(x_i|\mu_k, \Sigma_k)\right)\right).
    \end{equation}
For the minimization of the negative log-likelihood function, the EM algorithm (Expectation-Maximization algorithm) is commonly used to estimate the parameters of the GMM. The EM algorithm iteratively alternates between two steps: the E-step (Expectation step) and the M-step (Maximization step), until convergence to a local optimum. In the E-step, the probability that each sample belongs to each mixture component is estimated using the current parameters, and in the M-step, the model parameters are updated by maximizing a lower bound of the current likelihood function.

We utilize GMM to model and fit the distributions of these vectors, obtaining a Gaussian Mixture Model that describes these distributions. We save the mean, covariance matrix, and weight coefficient of each Gaussian component. In the next incremental phase, we generate samples that conform to the mixed Gaussian distribution, thus preserving knowledge. However, considering the issue of the number of feature subspaces, i.e., not every class necessarily requires $K$ Gaussian components for fitting, we achieve adaptive selection of the number of Gaussians during the GMM model fitting process. That is, to obtain the optimal number of feature subspaces for the feature distribution, we first calculate the closeness of sample $i$ to other samples in the same class:
\begin{equation}
    a(i) = \frac{1}{|C_i| - 1} \sum_{j \in C_i, j \neq i} d(i, j)^2,
\end{equation}
where $C_i$ denotes the set of samples in the same class as sample $i$, $|C_i|$ is the number of samples in set $C_i$, and $d(i, j)$ represents the Euclidean distance between sample $i$ and sample $j$.
We then calculate the separateness of sample $i$ from all samples in every other different class:
\begin{equation}
    b(i) = \min_{C_k \neq C_i} \frac{1}{|C_k|} \sum_{j \in C_k} d(i, j)^2.
\end{equation}
We obtain the silhouette coefficient of sample $i$:
\begin{equation}
    s(i) = \frac{b(i) - a(i)}{\max\{a(i), b(i)\}}.
\end{equation}
Finally, by averaging the silhouette coefficients of all samples, and comparing the average silhouette coefficients under different numbers of Gaussians, we can determine the optimal value of $K$.
Additionally, we need to consider that if there are no longer multiple feature subspaces within a class, meaning the true features may only need to be modeled and fitted with a multivariate Gaussian distribution, we set a threshold for the silhouette coefficient. Only when it exceeds this threshold do we believe that $K$ Gaussian components are needed; otherwise, only one Gaussian component is used.

Furthermore, if using a multivariate Gaussian distribution, we need to store the covariance matrix. For example, using ResNet18 \cite{he2016resnet} as the backbone network, we need to store a 512-dimensional class center vector and a $512 \times 512$ covariance matrix. This imposes a higher storage burden compared to prototype-based methods. To address this, we perform twice degradation of the covariance matrix, continuously aligning it for dimensionality reduction, while preserving the centers and weights information obtained from the GMM (as there is no storage burden). We then analyze this process, resulting in a more comprehensive solution. Details will be provided in the experimental section.

\begin{algorithm}

\caption{DMR Modeling and Fitting in Continual Learning}
\label{alg:gmm}
\begin{algorithmic}

\Procedure{Fitting Feature Distribution in Current-Task}{}
\State \textbf{Input:} Features set {$\{x_1, x_2, ..., x_n\}_{c_{i}}$} of class $c_{i}$, task\_id
\For{$c_{i}$ in task $\tau$}
    \State $\text{Silhouette Score Algorithm}(\{x_1, x_2, ..., x_n\}_{c_{i}})$
    \State $K^* = \text{adaptive} \_K( \text{silhouette scores})$
    \State $\{\mu_{c_i}, \Sigma_{c_i}, \pi_i\}_{c_i}^{K^*} = \text{GMM Fitting}(\{x_1, x_2, ..., x_n\}_{c_{i}}, K^*)$
    \State Store parameters: $\{\mu_i, \Sigma_i, \pi_i\}_{i}^{K^*}$
\EndFor
\State \Return Stored parameters
\State \hrule
\EndProcedure

\Procedure{Generate pseudo Features for Old-Task}{}
\State \textbf{Input:} Store parameters:$\{\mu_{c_i}, \Sigma_{c_i}, \pi_i\}_{c_i}^{K^*}$
\For{task 0, 1,..., $\tau - 1$}
  \For{$c_{i}$ in task}
    \State $\text{GMM Generation Algorithm}$ $( \{\mu_{c_i}, \Sigma_{c_i}, \pi_i\}_{c_i}^{K^*} )$
  \EndFor
\EndFor
\State \Return Generate pseudo Features
\EndProcedure

\end{algorithmic}

\end{algorithm}

\subsection{Avoiding Old-New Confusion}
\label{imfe}
\begin{figure}[t]
    \centering
    \includegraphics[width=\linewidth]{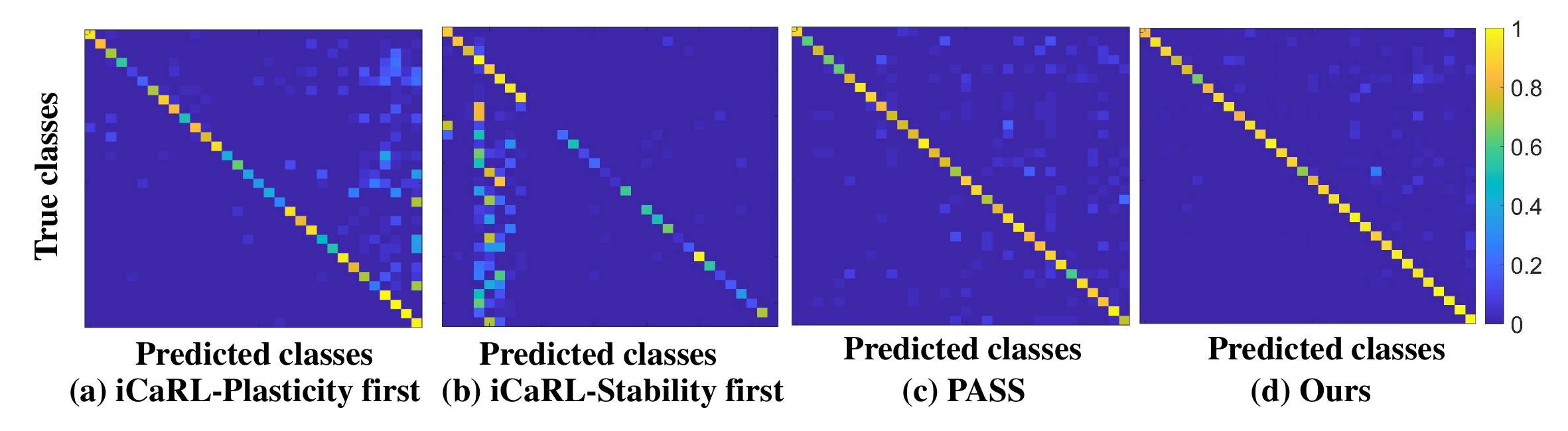}
    \caption{The relationship between confusion level and plasticity and stability in tasks old and new. By controlling the coefficient of distillation loss, the contrasts in (a) and (b) in the figure are both severely confused; pass better alleviates the dilemma of plasticity and stability, but there is still confusion between new and old classes; our method achieves better results by avoiding confusion between new and old.} 
    \label{cfm} 
\end{figure}
\begin{figure}[h]
    \centering
    \includegraphics[width=0.6\linewidth]{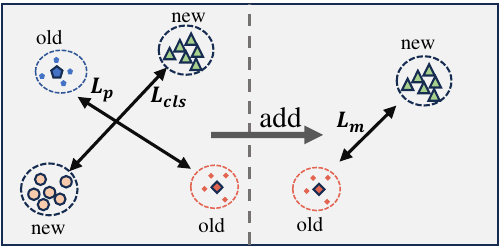}
    \caption{Unlike the two losses in previous prototype-based methods, we enhance the pseudo features using samples from the new task, increasing the discriminative ability between new and old tasks.} 
    \label{mix-proto} 
\end{figure}
Through the method in \ref{DMR}, we can effectively retain old knowledge, aiming to minimize confusion between old classes. However, as shown in Fig. \ref{cfm}, we should also enhance discriminability between new and old tasks. The representation of old knowledge by pseudo features is fixed in the embedding space. When training with new samples, confusion may arise, leading to classifier drift. Therefore, we aim to make the classifier more discriminative towards new and old knowledge. To achieve this, we enhance the prototype with features from new classes outside the old knowledge domain. 

In task $t+1$, the representation $e_{n}^{{c}_1}$ of the $n$-th sample from class $c_1$ in the current task is obtained. We then generate the pseudo features $\varphi _{p}^{c_2}$ for a specific old class $c_2$ based on \ref{DMR}. Subsequently, in the incremental feature space, we blend the obtained pseudo-prototype $\varphi _{p}^{c_2}$ of the old class with the feature $e_{n}^{c_1}$ of the new class in a linear manner, akin to the Mixup approach \cite{zhang2017mixup}. The training label is the hard label of the new class, which is utilized to distinguish between new and old classes.
\begin{equation}
    \varphi _m = \lambda \cdot e_{n}^{c_1} + \left ( 1-\lambda  \right ) \cdot  \varphi _{p}^{c_2},
\end{equation}
where $\lambda \in \left [0, 1 \right ]$ follows a Beta distribution.
In Fig. \ref{mix-proto}, the three types of data, $e_{n}^{c_1}$, $\varphi _{p}^{c_2}$, and $\varphi _m$, representing the new class, the old class, and the mixture of new and old classes, respectively, are input to the linear classifier. They are supervised by three corresponding loss functions, $L_{cls}$, $L_{P}$, and $L_{m}$, all computed using cross-entropy. 
In the feature space, $\varphi_{m}$, enhanced by new knowledge prototypes, lies between the features of the new and old classes. This approach mitigates overfitting of the classifier to the prototypes and enhances its discriminability between new and old classes.
In summary, the loss function is composed as:
\begin{equation}
    L = \xi \left ( L_{cls} + L_{p} \right )  + \left ( 1- \xi \right ) L_{m},
\end{equation}
where $\xi \in \left [ 0,1\right ]$ is a hyperparameter.

Additionally, we categorize classification errors into two types: confusion within tasks and confusion between new and old tasks.
Let $N$ represent the total number of samples in the new task, $O$ represent the total number of samples in the old task, $M_{\text{new}}$ represent the number of samples misclassified from the old task to the new task, and $M_{\text{old}}$ represent the number of samples misclassified from the new task to the old task. The confusion index between the tasks can be calculated as:
\begin{equation}
    C\_I = \frac{M_{\text{new}}}{N} + \frac{M_{\text{old}}}{O}.
\end{equation}
When accumulating the confusion across different incremental stages, you can sum the confusion values of each stage. Assuming there are $K$ incremental stages (tasks), with each stage having a confusion index of $C\_I_{\tau}$, the total confusion can be expressed as:
\begin{equation}
    C\_I_{total} = \sum_{\tau=1}^{K} C\_I_{\tau}.
\end{equation}

\subsection{Generalization Is Prerequisite.}
\label{3.B}
\begin{figure}[t]
    \centering
    \includegraphics[width=\linewidth]{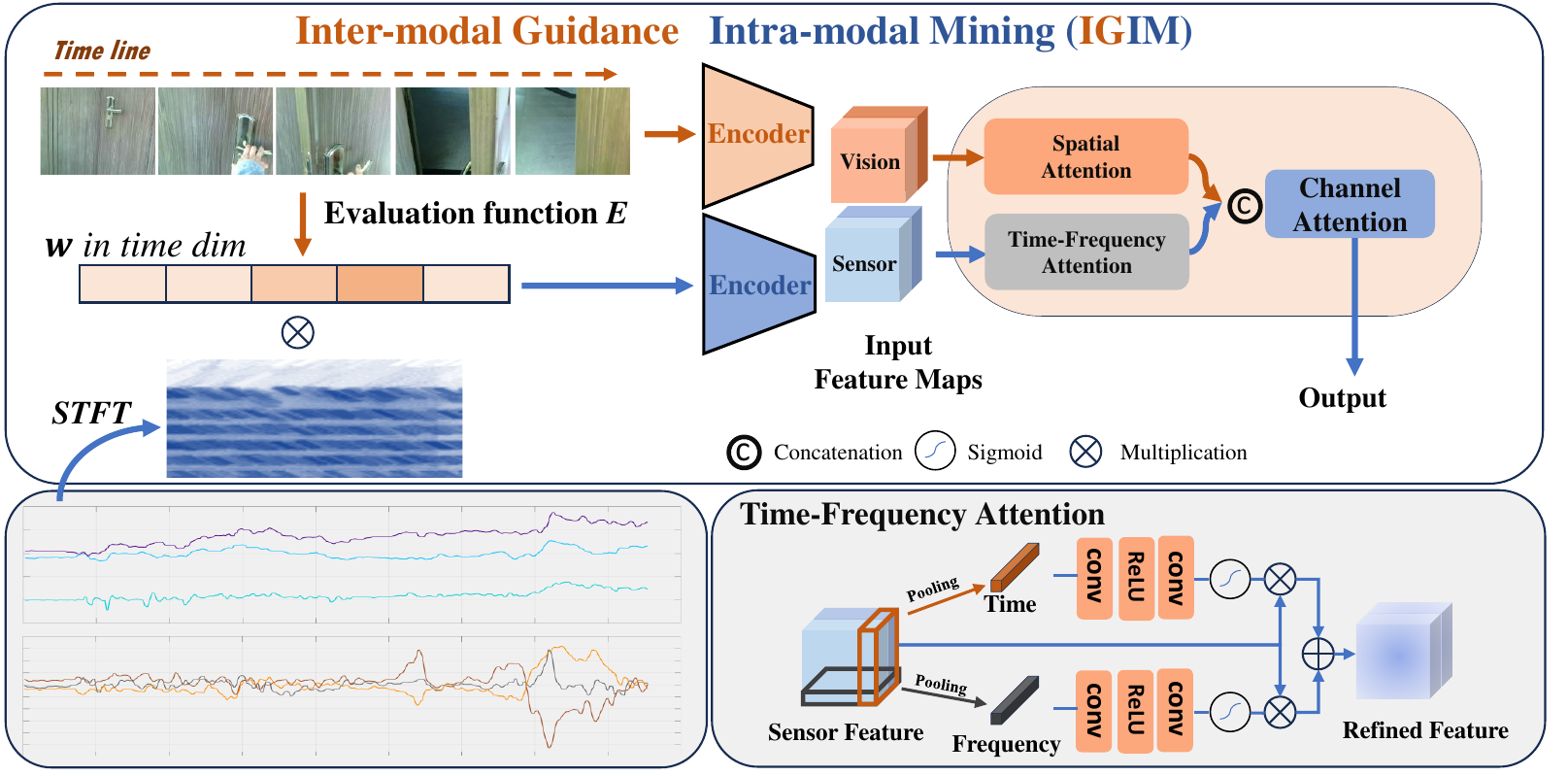}
    \caption{Detail architecture of IGIM. It aims to alleviate the imbalance of sensor modalities caused by the optimization deficiency of weak sensor modalities through enhancing information on the time and frequency dimensions.} 
    \label{mm-framework} 
    \vspace{-1em}
\end{figure}

\begin{figure}[t]
    \centering
    \includegraphics[width=0.7\linewidth]{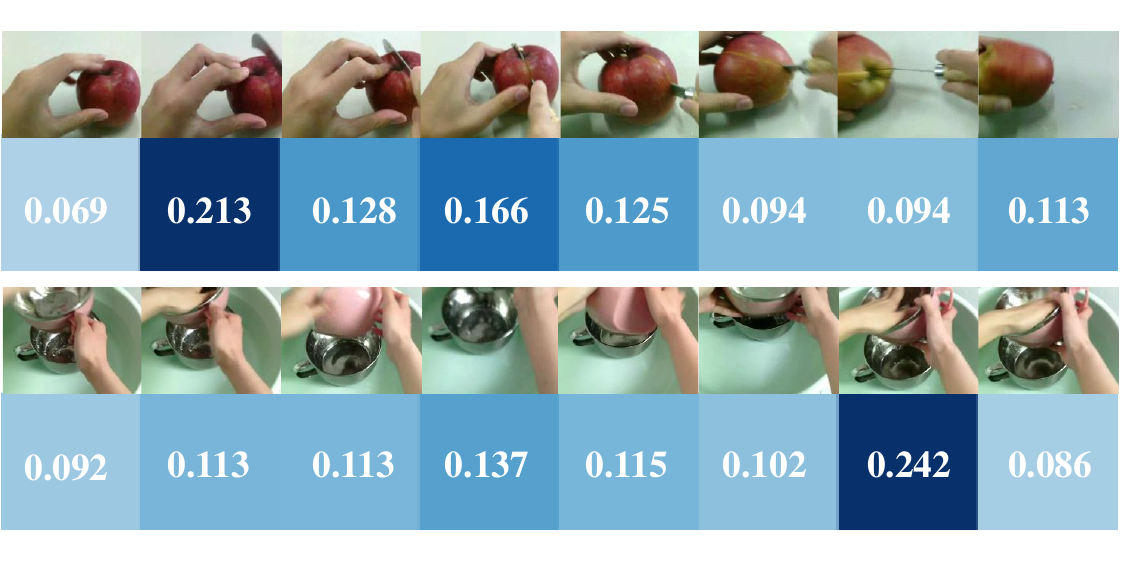}
    \caption{The visualization of time-guided weights in different classes. Along with their corresponding images, is presented. In the figure, darker colors indicate higher weights assigned to sensor modalities at the corresponding time.} 
    \label{time-weights} 
    \vspace{-1em}
\end{figure}

In continual learning or Multimodal Continual Learning, it is essential to ensure that the network performs well in each stage, which requires adaptability, the ability to adapt to new tasks. In existing research, it has been found that the balance of multimodal is one of the key factors affecting the performance of multimodal training. We also can see in Fig. \ref{balance}, the feature response difference between different modalities is very large, as a result, the distribution of old knowledge cannot be easily reproduced. Therefore, maintaining the balance between modalities is one of the sufficient conditions for multimodal continual learning.
In multimodal continual learning, the $\mathcal{D_\tau}=\left\{ x^{v}_{\tau,i},x^{s}_{\tau,i},y_{\tau,i}\right\}_{i=1}^{N_{\tau}}$ represents the input of $N_{\tau}$ paired visual modality data and sensor modality data $x^{v}_{\tau,i},x^{s}_{\tau,i}\in \mathbb{R}^{D}$ at stage $\tau$, with the corresponding label $y_{\tau,i} \in C_{\tau}$. We decouple the entire model into an encoder, an Intra-modal mining module, and a classifier:
\begin{equation}
    f\left( x^{v}_{\tau,i},x^{s}_{\tau,i} \right)=W^{\top} \mathcal{M}( \phi_{v}(x^{v}_{\tau,i}),\phi_{s}(x^{s}_{\tau,i})),
\end{equation}
where $\phi_{\ast}(\cdot):\mathbb{R}^{D} \to \mathbb{R}^{H\times W \times d},\mathcal{M}(\cdot) :\mathbb{R}^{H\times W \times d} \to \mathbb{R}^{d} , W \in \mathbb{R}^{d \times  |y_{\tau}|}$, and the encoder consists of two parallel single-modality encoders $\phi_{v}(\cdot)$ and $\phi_{s}(\cdot)$, outputting feature maps.
Firstly, as shown in Fig. \ref{mm-framework} inspired by transfer learning methods, since there is an imbalance between dominant and weak modalities, we transfer effective prior information obtained from dominant modalities to weak modalities, thereby alleviating the insufficient information in weak modalities caused by under-optimization. The input of the visual modality ${x^{v}_{\tau,i}}_{j}|_{j=1}^{k}$ contains $k$ frames extracted at intervals from the same video arranged in chronological order, which is aligned in time with the sensor signal. The images containing time information are inputted into the Evaluation function $E(\cdot)$. Inspired by \cite{wang2022dualprompt}, we directly use pretrained models as part of $E(\cdot)$. Through $E(\cdot)$, we obtain the weight information of the visual modality on the timeline and weight this information onto the temporal dimension of the spectrogram obtained from the sensor signal as the input of the sensor modality network:
\begin{equation}
\begin{aligned}
I_{S_{\tau,i}} &= w^{t} \cdot x^{s}_{\tau,i} = E(x^{v}_{\tau,i}) \cdot x^{s}_{\tau,i} \\
&= \frac{e^{\sum_{d} |p(\phi_{v}({x^{v}_{\tau,i}}_{j}))|/T}}{\sum_{j=1}^{8}e^{\sum_{d} | p(\phi_{v}({x^{v}_{\tau,i}}_{j}))| / T}} \cdot x^{s}_{\tau,i},
\end{aligned}
\end{equation}

\begin{equation}
I_{V_{\tau,i}} = x^{v}_{\tau,i},
\end{equation}
where $w^{t}$ denotes the information extracted from the visual modality over time, $p(\cdot)$ denotes pooling over the width and height of the feature map output by $\phi_{v}(\cdot)$. After obtaining $I_{V_{\tau,i}}$ and the time-guided $I_{S_{\tau,i}}$, we then perform intra-modal information mining. From this perspective, inspired by \cite{cbam}, as shown in Fig. \ref{time-weights}, we explore and enhance information in the sensor network input $I_{S_{\tau,i}}$ in both the time and frequency domains. The feature map of the sensor modality contains three dimensions: channel, time, and frequency. Same as our previous method \cite{csx}, in Time-Frequency attention, we perform pooling in both the time and frequency dimensions, then feed them into the small net shown in the Fig. \ref{mm-framework}. Finally, we add them back to the original input feature map in the time and channel dimensions, forming the sensor modality's information miner. The information that the visual modality should focus on comes from spatial aspects, so we refer to \cite{cbam}. After obtaining the enhanced feature map, we simply concatenate them and input them into the channel attetion for fusion.

\section{Experiments}

\subsection{Datasets}
\label{datasets}
We utilized two commonly used datasets in continual learning, CIFAR100 and ImageNet100. Additionally, considering the challenges of modal imbalance in our multimodal applications, which affect the generalization capability of multimodal models, we extended our study to the UESTC-MMEA-CL dataset for multimodal behavior recognition. When the generalization ability of multimodal models is ensured, we are willing to extend our continual learning methods through experimental validation.

\textbf{CIFAR100} \cite{krizhevsky2009learning(cifar)}is a dataset widely used for class-incremental learning. It contains 100 classes, each with 600 RGB images. For each class, 500 images are used for training and 100 for testing.

\textbf{ImageNet100} \cite{Rebuffi17(icarl)} is a subset of ImageNet-1K \cite{russakovsky2015imagenet}, consisting of 100 classes randomly selected after shuffling the original 1000 classes. Each class in ImageNet100 contains an average of 1300 training samples and 50 test samples. The ImageNet dataset is widely recognized in the field of computer vision and is extensively used in continual learning research.

\textbf{UESTC-MMEA-CL} \cite{xu2023towards} is a multimodal activity recognition dataset designed for continual learning. It contains 32 daily behavior classes, with approximately 200 samples per class. The training and testing ratio is 7:3, and each sample consists of temporally aligned multimodal segments, including video segments, accelerometer signal segments, and gyroscope signal segments. Same with He et al. \cite{he2024continual}, we sample 8 RGB images from each video segment and obtain optical flow data generated from the videos, and also perform short-time Fourier transform on the sensor signals to obtain time-frequency spectrograms, which are then input into the model.

\subsection{Implementation Details}
\label{Implementation Details}
 We need to compare our work with different works on different datasets. Since the tasks on each dataset are not exactly the same, the model structures and task settings for each dataset are not exactly the same. However, we can ensure a fair comparison to highlight the effectiveness of our work.

\textbf{Task Settings}:In the CIFAR100 dataset, consistent with the prototype-based continual learning method setting, we divide the 100 classes into $T=5$, $T=10$, $T=20$, and $T=60$, specifically set as  $50 + 10 \times 5$, $50 + 5 \times 10$, and $40 + 3 \times 20$  ( $50 + 10 \times 5$ means the training process is divided into one base phase and 5 incremental phases, with 50 classes in the base phase and 10 classes in each incremental phase). On the ImageNet100 dataset, it is also divided into $T=5$, $T=10$ and $T=20$, the detailed settings are also consistent with those used on CIFAR100. In the UESTC-MMEA-CL dataset, we shuffle the 32 classes with a fixed random seed, consistent with \cite{xu2023towards}. We divide them into three cases: $T=16$, $T=8$, and $T=4$, specifically set as  $2 + 2 \times 15$, $4 + 4 \times 7$, and $8 + 8 \times 3$. 

\textbf{Architecture and train details:} On the CIFAR100 and ImageNet100 dataset, we use ResNet18 for training, with parameters set the same as in FeTrIL \cite{petit2023fetril}. In the base phase, the learning rate is 0.1 for 200 iterations, and in the incremental phase, it is 0.05 for 60 iterations. All the models are built based on pytorch framework, and the method implementation and experiment are mainly implemented based on PyCIL \cite{zhou2021pycil}.
In the research on the UESTC-MMEA-CL dataset, the encoder is BN-Inception pre-trained on KICHENS-100, and the entire backbone network structure is consistent with AID \cite{csx}. The training details are also consistent with it, training 20 epochs for each stage, using a momentum of 0.9, a batch size of 8, a learning rate of 0.001, and a learning rate decay at the 10th epoch. We use the SGD optimizer for training.

\subsection{Analysis on DMR and DMR-L Method}
\label{analysis on DMR}
\begin{figure}[h]
    \centering
    \includegraphics[width=\linewidth]{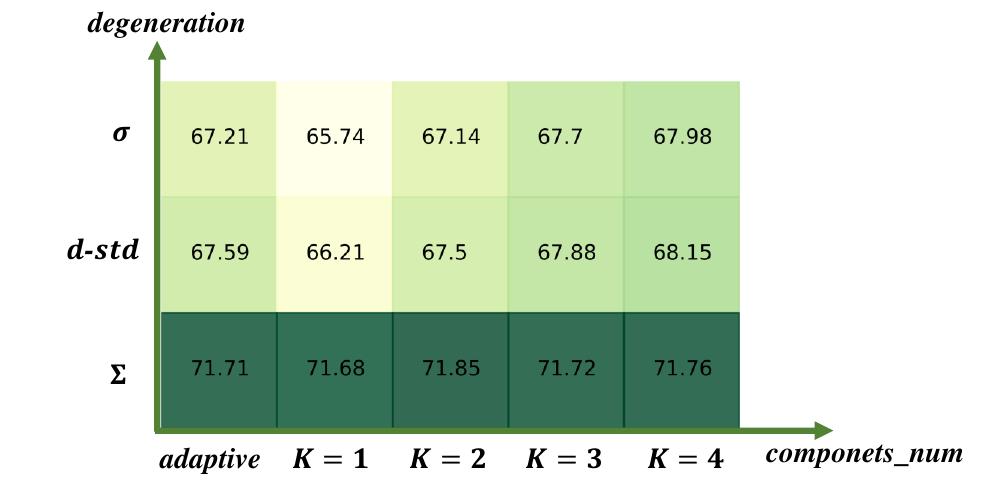}
    \vspace{0em}
    \caption{On CIFAR100 dataset, different Gaussian components and two sets of information degradation are used to carry out memory recall of the old knowledge at the distribution level, and the final joint result is represented by the average accuracy.} 
    \label{k_sigma} 
\end{figure}
In \ref{DMR}, we encountered two issues when modeling withGMM: determining the number of Gaussians for modeling and the size of the covariance matrix. We also proposed corresponding solutions. As shown in the Fig. \ref{k_sigma}, we conducted extensive experiments on different numbers of Gaussians and the degradation of information in the covariance matrix.

Firstly, by observing the Fig. \ref{k_sigma} horizontally, we noticed that as $K$ increases from 1, the average accuracy generally rises. However, the significant improvement occurs mainly between 1 and 2, with little improvement afterward and even some fluctuations, indicating a saturation phenomenon. Although the mean value of $K$ for each class using the adaptive method is less than 2. In the UESTC-MMEA-CL dataset and CIFAR100 dataset, the threshold values we selected were both 0.1, and the $\bar{K}$ obtained were 1.47 and 1.35, respectively the effect is better than $K$=2. This phenomenon is interesting; we attribute it to the anisotropy in the feature distribution during the incremental process. Some classes may contain a single feature subspace, while others contain multiple subspaces. In such cases, using a fixed $K$ value cannot meet the modeling requirements of all classes.

Looking at the Fig. \ref{k_sigma} vertically, as expected, the average accuracy decreases as the information in the covariance matrix degrades. However, it is interesting to note that using the vector $d$-$std$ composed of the square roots of the diagonal elements of the covariance matrix does not significantly outperform using the trace of the covariance matrix divided by the dimensionality and then take the square root to obtain the standard variance $\sigma$. The former is only slightly higher, around 0.3, but it uses floating-point numbers of a higher order of magnitude. Of course, modeling with $\Sigma$ to generate pseudo features in the new stage far outperforms the degraded method. Mathematically, this is because there is correlation among the dimensions of the feature vectors, not independence. Consequently, after one or two degradations, the loss of correlation information leads to similar average accuracy. Finally, considering both aspects, we define the method corresponding to the adaptive axis using the covariance matrix $\Sigma$ as DMR and the method using $\sigma$ as DMR-Lite. DMR method maintains very high accuracy, while DMR-Lite sacrifices performance to alleviate the burden of storage but still surpasses the SOTA. 

In addition to the above, we also need to consider: \textbf{Are the pseudo features generated in the new stage consistent with the features of real samples?}
We can use kernel functions and Maximum Mean Discrepancy (MMD) to assess the similarity between the clusters formed by the two sets of feature vectors. MMD is an unbiased estimate that measures the distance between two kernel matrices. The calculation is as follows:
\begin{equation}
\begin{aligned}
     \text{MMD}^2 & = \frac{1}{n(n-1)} \sum_{i=1}^{n} \sum_{j\neq i}^{n} k(x_i, x_j) - \frac{2}{n m} \sum_{i=1}^{n} \sum_{j=1}^{m} k(x_i, y_j) \\
     &+ \frac{1}{m(m-1)} \sum_{i=1}^{m} \sum_{j\neq i}^{m} k(y_i, y_j).
\end{aligned}
\end{equation}
Here, $k(\cdot, \cdot)$ is the kernel function, and $n$ and $m$ are the numbers of samples in the two clusters. The MMD value can be used to evaluate the similarity between the two clusters. A smaller MMD value indicates greater similarity between the clusters, while a larger MMD value indicates less similarity.
\begin{figure}[htbp]
        \centering
        \includegraphics[width=0.9\linewidth]{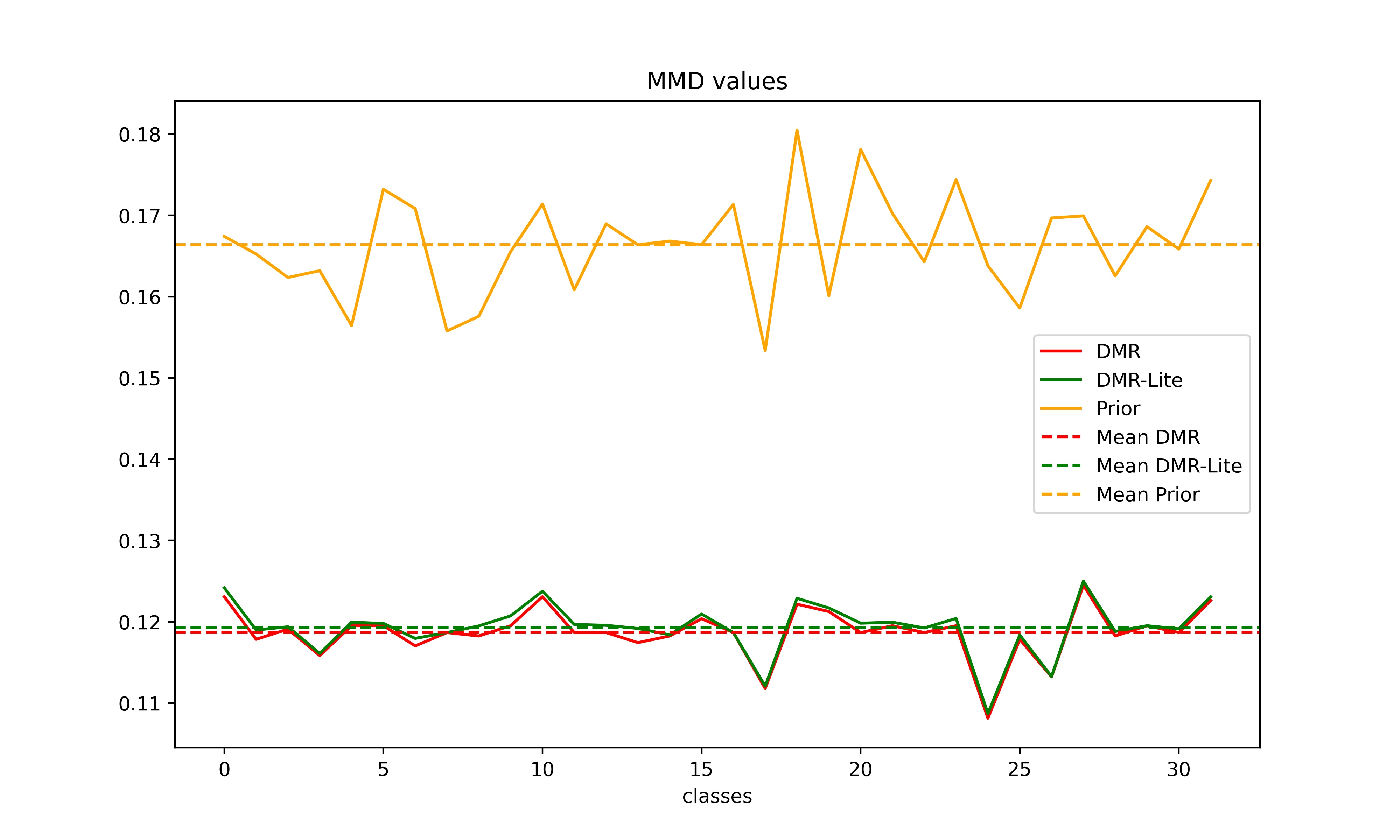}
        \vspace{-1em}
        \caption{\textbf{The degree of retention of old knowledge by different methods is compared.} The larger the MMD value, the greater the difference between the distribution of the ``recall'' simulation and the distribution of the real sample in the embedding space.} 
        \vspace{-1em}
        \label{mmd} 
\end{figure}

By recalling the data of the class centers and other information, we contrast the old knowledge with the distribution of the feature space of real samples. Using the MMD evaluation method proposed above, in the DMR method, we generate the distribution in the old class embedding space through the preserved old class centers and covariance matrices. In the DMR-Lite method, we simulate the distribution of old knowledge by using the preserved old class cluster centers and cluster mean square deviations, and through linear interpolation. In the Prior method, we simulate old knowledge using the most basic class centers and mean square deviations. The simulated distributions of old knowledge generated by the three methods all utilize the Gaussian kernel function to calculate the MMD value between the simulated distribution and the real distribution. As shown in the Fig. \ref{mmd}, we can see that the distributions of Prior, DMR-Lite, and DMR with real samples in the embedding space become more and more similar. The DMR-Lite method saves storage burden, but the effect of preserving old knowledge has not significantly decreased. 

\subsection{Analysis on IMFE Method}
\begin{figure}[htbp]
        \centering
        \includegraphics[width=0.8\linewidth]{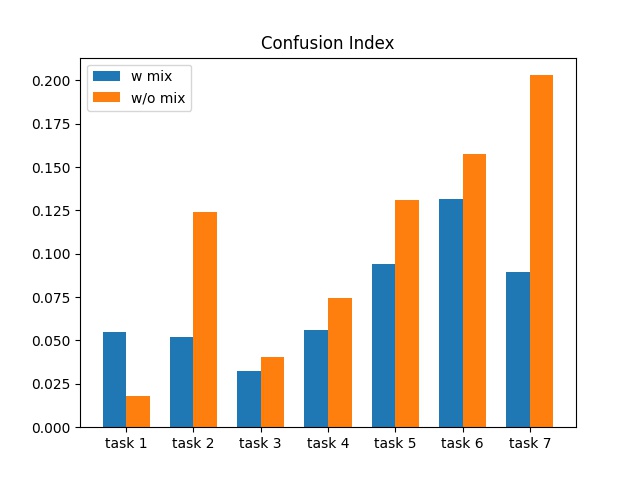}
        \vspace{-1em}
        \caption{Confusion between old and new tasks, a visual display of the Confusion-index.} 
        \vspace{-1em}
        \label{ci} 
\end{figure}

After preserving old knowledge, we also need to pay attention to avoiding confusion when learning new knowledge. We evaluate the confusion ability through a simple index called the Confusion Index, the explanation of which is in \ref{imfe}. In simple terms, it is the sum of the proportions of samples misclassified as new tasks in the old task and samples misclassified as old tasks in the new task. From the Fig. \ref{ci}, we can see that after performing mixup enhancement between the new and old classes, the degree of confusion has decreased. This indicates a reduction in the confusion between the new and old tasks.

\subsection{Comparison with Benchmarks}
\begin{figure*}[t]
    \centering
    \includegraphics[width=0.8\textwidth]{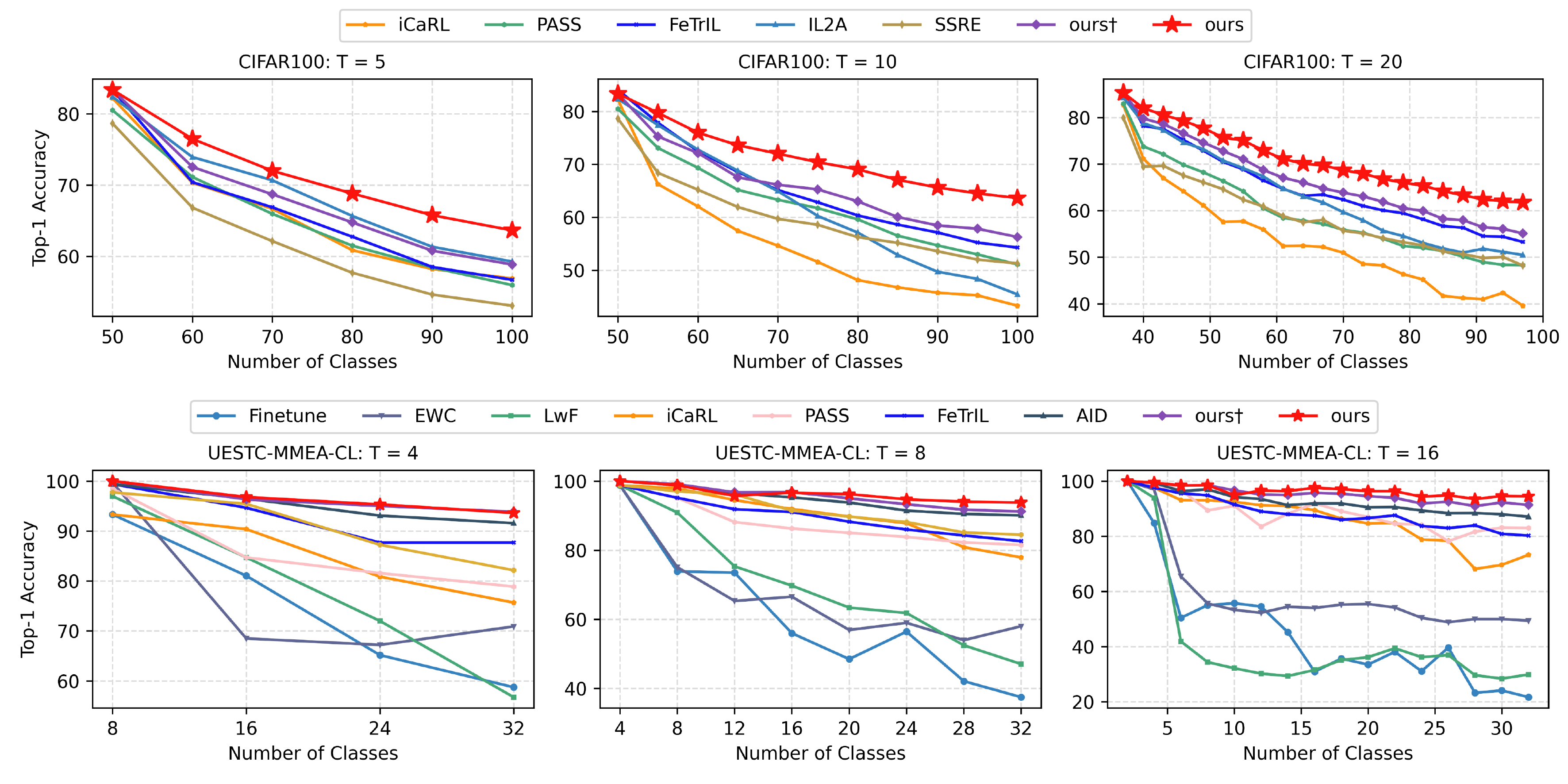}
    \caption{The Top-1 accuracy of different methods about different task settings on UESTC-MMEA-CL and CIFAR100 datasets, our$\dagger$ means using the DMR-L submethod.} 
    \label{forgetting} 
\end{figure*}
\textbf{Comparison Methods.}
To demonstrate the effectiveness of our method, we conducted extensive experiments on the aforementioned different datasets using the corresponding task settings. The comparison methods include rehearsal-based and prototype-based methods such as iCaRL \cite{Rebuffi17(icarl)}, LUCIR \cite{LUCIR}, PASS \cite{zhu2021prototype(pass)}, IL2A \cite{zhu2021class(IL2A)} and FeTrIL \cite{petit2023fetril}, as well as parameter regularization methods like EWC \cite{Kirkpatrick17(ewc)} and LwF \cite{li2017learning(LWF)}, and dynamic network methods like FOSTER \cite{wang2022foster}, SDC \cite{SDC}, DeeSIL \cite{belouadah2018deesil} and SSRE \cite{zhu2022self(SSRE)}. In our experiments, our method consistently outperformed the other methods and maintained its superiority throughout the entire incremental process.

\textbf{Evaluation Protocol.}
Similar to \cite{zhou2023class, limit}, we define the top-1 accuracy at the end of each stage as $\mathcal{A}_{\tau}$. After completing the training, we obtain the final stage accuracy $\mathcal{A}_{T}$. The average accuracy $\bar{\mathcal{A}} = \frac{1}{T} \sum_{\tau = 1}^{T}\mathcal{A}_{\tau}$ is calculated by averaging the accuracy at each stage. Additionally, we use the performance dropping rate (PD), i.e., $PD = \mathcal{A}_{0} - \mathcal{A}_{\tau}$, to accurately describe the degree of mitigation of catastrophic forgetting. Higher average accuracy and last accuracy, as well as lower PD, indicate less forgetting.

In Tables \ref{TABLE_1}, \ref{TABLE_2} and \ref{TABLE_3}, we present the experimental results on the CIFAR100, ImageNet100 and UESTC-MMEA-CL datasets. From the Fig. \ref{forgetting}, it can be seen that our method outperforms the SOTA methods on these datasets. Specifically, we achieve a significant improvement over the original method on the UESTC-MMEA-CL dataset. Since the DMR-L, DMR, and IMFE methods do not involve multi-modal aspects, it can be considered that the relevant experiments on CIFAR100 and ImageNet100 are consistent with solving the multi-modal imbalance problem on the UESTC-MMEA-CL multi-modal dataset. Therefore, experiments on the DMR and IMFE methods can be conducted on both datasets. We define the ``ours'' method in the CIFAR100 and ImageNet100 datasets experiments as the combination of the DMR-L or DMR method with the IMFE method. Regarding the results on the CIFAR100 dataset, for methods without replay, we can surpass the baseline by more than 5\%. Furthermore, by observing different task settings, we find that previous methods generally exhibit the same trend: the more incremental stages are set, the faster the performance declines. However, our method mainly relies on distributing-level memory recall, thus maintaining good performance in more task stages, rather than experiencing a significant decline. For example, the FeTrIL method enriches the features of old classes by judging the similarity between the centers of new and old classes and shifting the distribution of new classes to enrich the features of old classes. This method simulates the old distribution better than storing class centers and variance vectors but fails to consider the significant differences between distributions of different classes, leading to confusion in classification boundaries, especially when there are many incremental task settings, which exacerbates this issue. As for storing samples and distillation-based methods, they aim to preserve old knowledge by saving old samples or maintaining old models, but the constraints at the input and output levels may not allow for the complete development of the uninterpretable black box in the expected direction. In addition, we observe that the DMR-L method can maintain excellent performance while saving storage burden, similarly achieving higher performance in maintaining old knowledge than other methods. In conclusion, our methods all exhibit higher performance improvement compared to current methods.

\begin{table*}[t]
\vspace{0em}
\caption{Average accuracy $\bar{\mathcal{A}}$, last accuracy $\mathcal{A}_{T}$ and performance dropping rate $PD$ performance comparison on CIFAR100. The symbol $\dagger$ indicates the use of DMR-L, in contrast, in Ours, the DMR method is employed without the symbol. }
\vspace{0em}
\centering
\tiny
\renewcommand\arraystretch{1.2}
\resizebox{0.9\textwidth}{28mm}{
\setlength{\tabcolsep}{2.8mm}
\begin{tabular}{l ccccccccccc}
\toprule[0.7pt]
\multirow{2}{*}{Method}    
&\multicolumn{3}{c}{$T=5$} &\multicolumn{3}{c}{$T=10$}  &\multicolumn{3}{c}{$T=20$}\\ 

\cline{2-10}
 & \multicolumn{1}{c}{$\bar{\mathcal{A}}$} & \multicolumn{1}{c}{$\mathcal{A}_{T}$} & \multicolumn{1}{c}{$PD$} 
 & \multicolumn{1}{c}{$\bar{\mathcal{A}}$} & \multicolumn{1}{c}{$\mathcal{A}_{T}$} & \multicolumn{1}{c}{$PD$} 
 & \multicolumn{1}{c}{$\bar{\mathcal{A}}$} & \multicolumn{1}{c}{$\mathcal{A}_{T}$} & \multicolumn{1}{c}{$PD$}\\
 
 \hline
Finetune                     & 23.46 & 8.87  & 73.35    & 13.40 & 4.96  & 76.62     & 8.48  & 3.34  & 79.38 \\
EWC \cite{Kirkpatrick17(ewc)}      & 24.16 & 10.03 & 72.11    & 13.73 & 5.18  & 76.96     & 8.74  & 3.20  & 79.00 \\
LwF \cite{li2017learning(LWF)}     & 38.20 & 24.13 & 57.97    & 30.22 & 16.74 & 65.36     & 22.40 & 10.42 & 72.18 \\
iCaRL \cite{Rebuffi17(icarl)}    & 65.86 & 56.90 & 25.30    & 54.85 & 43.29 & 38.93     & 53.32 & 39.59 & 43.13 \\
PASS \cite{zhu2021prototype(pass)}  & 65.58 & 55.96 & 24.54    & 62.55 & 51.09 & 29.41     & 59.43 & 48.30 & 34.68 \\ 
IL2A \cite{wang2022foster}    & 68.87 & 59.27 & 23.11    & 61.82 & 45.41 & 36.97     & 62.96 & 50.48 & 33.82 \\
SSRE \cite{wang2022foster}    & 62.17 & 53.08 & 25.56    & 60.07 & 51.28 & 27.36     & 58.83 & 48.21 & 31.74 \\
FeTrIL \cite{petit2023fetril} & 66.43 & 56.7 & 25.15    & 65.15 & 54.27 & 25.96     & 64.91 & 53.27 & 32.68 \\
\hline 
Ours$\dagger$ &\textbf{68.17} & \textbf{58.86} &24.5  &\textbf{65.95} &\textbf{56.27} &27.09 &\textbf{66.60} & \textbf{55.11} &30.21 \\
Ours &\textbf{71.68} & \textbf{63.8} &19.56  &\textbf{71.52} &\textbf{63.97} &19.39 &\textbf{70.86} & \textbf{61.73} &23.59 \\
\bottomrule[0.7pt]
\end{tabular}
}
\label{TABLE_1}
\end{table*}

\begin{table}[t]
\vspace{0em}
\caption{Average accuracy $\bar{\mathcal{A}}$ on ImageNet100. }
\vspace{0em}
\centering
\scriptsize
\renewcommand\arraystretch{1}
\resizebox{0.8\linewidth}{!}{
\begin{tabular}{l ccc}
\toprule[0.7pt]
\multirow{1}{*}{Method}    
&\multicolumn{1}{c}{$T=5$} &\multicolumn{1}{c}{$T=10$}  &\multicolumn{1}{c}{$T=20$}\\ 
\cline{2-4}
 
 \hline
EWC \cite{Kirkpatrick17(ewc)} & - & 20.4 & -  \\
LwF-MC \cite{li2017learning(LWF)}   & - & 31.2 & -  \\
DeeSIL \cite{belouadah2018deesil}    & 67.9 & 60.1 & 50.5 \\
LUCIR \cite{LUCIR}& 56.8 & 41.4 & 28.5 \\
SDC \cite{SDC} & - & 61.2 & - \\
PASS \cite{zhu2021prototype(pass)}  & 64.4 & 61.8 & 51.3    \\ 
SSRE \cite{zhu2022self(SSRE)}    & - & 67.7 & -   \\
FeTrIL \cite{petit2023fetril} & 72.2 & 71.2 & 67.1  \\
\hline 
Ours$\dagger$ &\textbf{72.5} & \textbf{71.7} & \textbf{67.6}  \\
Ours &\textbf{76.0} & \textbf{75.8} &\textbf{72.5}   \\
\bottomrule[0.7pt]
\end{tabular}
}
\label{TABLE_2}
\end{table}

\begin{table*}[t]
\vspace{0em}
\caption{Average accuracy $\bar{\mathcal{A}}$, last accuracy $\mathcal{A}_{T}$ and performance dropping rate $PD$ performance comparison on UESTC-MMEA-CL.}
\vspace{0em}
\centering
\tiny
\renewcommand\arraystretch{1.2}
\resizebox{0.9\textwidth}{28mm}{
\setlength{\tabcolsep}{2.8mm}
\begin{tabular}{l ccccccccccc}
\toprule[0.7pt]
\multirow{2}{*}{Method}    &\multicolumn{3}{c}{$T=16$} &\multicolumn{3}{c}{$T=8$}  &\multicolumn{3}{c}{$T=4$}\\ 
\cline{2-10}
 
 & \multicolumn{1}{c}{$\bar{\mathcal{A}}$} & \multicolumn{1}{c}{$\mathcal{A}_{T}$} & \multicolumn{1}{c}{$PD$} & \multicolumn{1}{c}{$\bar{\mathcal{A}}$} & \multicolumn{1}{c}{$\mathcal{A}_{T}$} & \multicolumn{1}{c}{$PD$} & \multicolumn{1}{c}{$\bar{\mathcal{A}}$} & \multicolumn{1}{c}{$\mathcal{A}_{T}$} & \multicolumn{1}{c}{$PD$} \\ 
 \hline
Finetune                     & 45.20 & 21.66 &78.34     & 60.81 & 37.46 &61.32      & 74.57 & 58.74 &34.57 \\
EWC \cite{Kirkpatrick17(ewc)}      & 59.08 & 49.39 &50.61     & 66.68 & 57.98 &40.80      & 76.58 & 70.90 &28.80 \\
LwF \cite{li2017learning(LWF)}     & 41.55 & 29.86 &70.14     & 69.92 & 47.04 &51.74      & 77.61 & 56.76 &40.20 \\
iCaRL \cite{Rebuffi17(icarl)}    & 85.22 & 74.77 &25.23     & 90.56 & 84.05 &14.73      & 84.76 & 77.58 &15.73 \\
PASS \cite{zhu2021prototype(pass)}  & 88.13 & 82.98 &17.02     & 87.61 & 81.50 &17.28      & 91.95 & 85.41 &19.60 \\
FeTrIL \cite{petit2023fetril} & 88.47 & 80.24 &19.76     & 89.73 & 82.60 &16.18      & 92.35 & 87.69 &11.70 \\
FOSTER \cite{wang2022foster}  & -     & -     &-         & 91.35 & 84.45 &14.33      & 90.64 & 82.15 &15.59 \\
AID \cite{csx}  & 92.35 & 87.08 &12.92     & 94.54 & 90.05 &9.95       & 95.13 & 91.57 &7.82 \\
\hline 
Ours$\dagger$ & 95.08 & 91.34 &8.66     & 95.39 & 91.57 & 8.43       & 96.29 & 93.84 & 6.16 \\
Ours &\textbf{96.43} &\textbf{94.45} &5.55 
&\textbf{96.23} & \textbf{93.77} &6.23 
&\textbf{96.43} & \textbf{93.62} &6.38 \\
\bottomrule[0.7pt]
\end{tabular}
}
\label{TABLE_3}
\end{table*}

\subsection{Ablation Study}
\begin{figure}[h]
        \centering
        \includegraphics[width=0.9\linewidth]{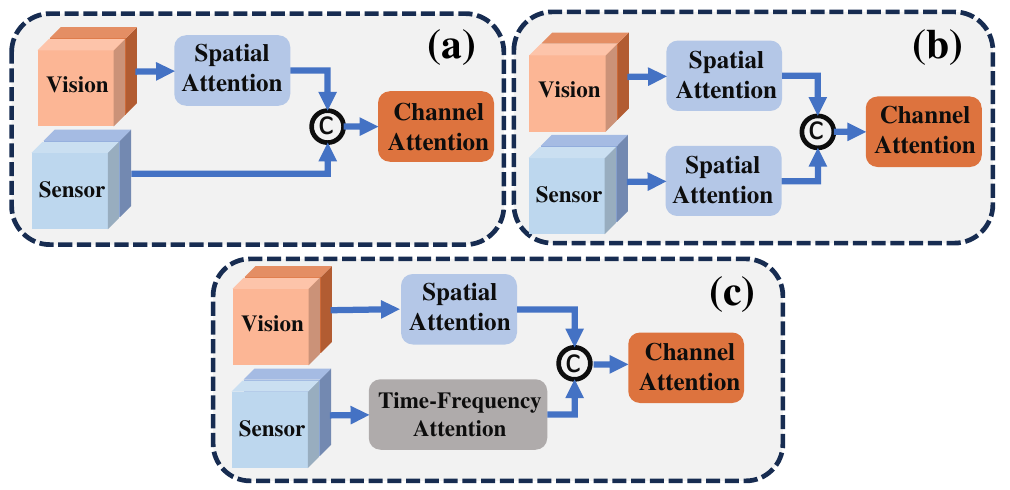}
        \vspace{0em}
        \caption{Proof of the effectiveness of our proposed Time-Frequency mining module by changing the composition of Intra-modal mining module.} 
        \label{Ablation Study on Time-Frequency mining module} 
\end{figure}

\textbf{Ablation on DMR-Lite and DMR.}
Since the DMR and DMR-Lite method does not involve improving the multimodal generalization ability, we conducted ablation experiments on two datasets. The final results are shown in the table below. We define the model that maintains old knowledge by adding class centers and a mean square deviation to the finetuned model as ``base''. We define the model that replaces the mean square deviation with a vector obtained by replacing it with the standard deviation on each dimension as ``base + $d$-$std$''. ``Base + DMR'' or ``base + DMR-L'' refers to the use of our method for replacement. As in session 3.3, in addition to accuracy, we should focus on the memory burden of different prototype maintenance methods for old knowledge. We use per-class memory burden (PMD) to evaluate, measured in a floating-point number. Where $d$ represents the feature dimension output by the encoder, such as $d=512$ in ResNet18 \cite{he2016resnet}, divided into two parts, ``$d+d$'' represents the class center vector composed of $d$ float-type numbers and the standard deviation vector composed of $d$ numbers, and ``$d+d^{2}$'' represents that in addition to a d-dimensional class center vector, a $d^2$ covariance matrix needs to be stored, the $\bar{K}$ in the table, as described in \ref{analysis on DMR}, takes 1.47 and 1.35 on the two datasets.

From the table, it can be seen that our method's ability to maintain old knowledge is much higher than the original prototype-based method. However, for DMR, the storage burden is significant, but it is acceptable for current computer hardware. In summary, the DMR-L method has made concessions in terms of storage burden while still maintaining the ability to preserve old knowledge.
\begin{table}[t]
\caption{Ablation experiments of DMR and DMR-L methods on UESTC-MMEA-CL and CIFAR100 datasets, and analysis of prototype-based storage burden.}
\centering
\renewcommand\arraystretch{1.3}
\resizebox{0.9\linewidth}{!}{
\begin{tabular}{lccccc}
\toprule[0.7pt]
\multirow{2}{*}{}  
&\multicolumn{2}{c}{U-M-CL $T = 8$} 
&\multicolumn{2}{c}{CIFAR $T = 5$} 
&\multirow{2}{*}{$\mathcal{PMD}$} \\ 
\cline{2-5} 
& \multicolumn{1}{c}{$\bar{\mathcal{A}}$} 
& \multicolumn{1}{c}{$\mathcal{A}_{T}$} 
& \multicolumn{1}{c}{$\bar{\mathcal{A}}$} 
& \multicolumn{1}{c}{$\mathcal{A}_{T}$} \\ 
\hline
base            &93.34  &86.63 &65.74 &55.98 &$d+1$  \\ 
base + $d$-$std$    &93.56  &88.53 &66.21 &56.53 &$d+d$   \\
base + DMR-L    &95.47  &91.19 &67.94 &58.98 &$\bar{K} (d+1)$ \\
base + DMR      &\textbf{95.91} & \textbf{93.24} &\textbf{71.68} &\textbf{63.65} &$\bar{K} (d+d^{2})$ \\
\bottomrule[0.7pt]
\end{tabular}
}
\label{ablation-dmr}
\end{table}

\textbf{Ablation on IMFE.}
Similarly, we conducted ablation experiments on the IMFE method with the PASS \cite{zhu2021prototype(pass)} and FeTrIL \cite{petit2023fetril} methods, both of which are prototype-based. The experimental results are shown in Table \ref{imfe1}.It can be observed that for prototype-based methods that maintain old knowledge in the feature space, the IMFE method enhances the distinctiveness between old and new knowledge, leading to a better classification performance compared to not using the IMFE method.

Furthermore, it can be observed from Table \ref{imfe2} that we conducted ablation experiments on the IMFE method within our own method. Similarly, it can be seen that using IMFE can improve the discriminative ability between new and old knowledge. However, compared to Table \ref{imfe1}, the improvement is not very significant. We believe this is because the DMR method is very effective in maintaining old knowledge, thus reducing the urgency for the IMFE method to improve the separability between new and old knowledge.

\begin{table}[ht]
\caption{The IMFE method was applied to the prototype based methods PASS \cite{zhu2021prototype(pass)}, FeTrIL \cite{petit2023fetril} and base for ablation experiments at UESTC-MMEA-CL dataset}
\vspace{0.5em}
\centering
\renewcommand\arraystretch{1.3}
\resizebox{0.9\linewidth}{!}{
\begin{tabular}{lcccc}
\toprule[0.7pt]
\multirow{2}{*}{Method}    
&\multicolumn{2}{c}{$T = 8$}  
&\multicolumn{2}{c}{$T = 4$} \\ \cline{2-5} 

& \multicolumn{1}{c}{$\bar{\mathcal{A}}$} 
& \multicolumn{1}{c}{$\mathcal{A}_{T}$} 
& \multicolumn{1}{c}{$\bar{\mathcal{A}}$} 
& \multicolumn{1}{c}{$\mathcal{A}_{T}$} \\  
\hline
PASS \cite{zhu2021prototype(pass)}  &87.61 &81.50 &91.95 &85.41 \\
PASS \cite{zhu2021prototype(pass)} $w/ \, IMFE$  &91.03  & 84.42   &92.24 &88.22\\
FeTrIL \cite{petit2023fetril}   &89.73 & 82.60 &92.35 &87.69\\
FeTrIL \cite{petit2023fetril} $w/ \, IMFE$  & 89.74 & 81.23 &93.17 &87.99\\
base   &94.07 & 88.68 &94.03 &91.11 \\
base   $w/ \, IMFE$  &\textbf{94.54} & \textbf{90.05} &\textbf{95.13} &\textbf{91.57}\\
\bottomrule[0.7pt]
\end{tabular}
}
\label{imfe1}
\end{table}

\begin{table}[ht]
\caption{Ablation experiments were performed on the IMFE submethod on the CIFAR100 dataset on our proposed method.}
\vspace{0.5em}
\centering
\renewcommand\arraystretch{1.3}
\resizebox{0.9\linewidth}{!}{
\begin{tabular}{lcccc}
\toprule[0.7pt]
\multirow{2}{*}{Method}    
&\multicolumn{2}{c}{$T = 5$}  
&\multicolumn{2}{c}{$T = 10$} \\ 
\cline{2-5} 

& \multicolumn{1}{c}{$\bar{\mathcal{A}}$} 
& \multicolumn{1}{c}{$\mathcal{A}_{T}$} 
& \multicolumn{1}{c}{$\bar{\mathcal{A}}$} 
& \multicolumn{1}{c}{$\mathcal{A}_{T}$} \\  
\hline
base + DMR-L   &67.94  &58.98   &65.36  &55.75  \\
base + DMR     &71.68  &63.65   &71.38  &63.63  \\
base + DMR-L $w/ \, IMFE$ &\textbf{68.17} & \textbf{58.86} &\textbf{65.95} &\textbf{56.27}\\
base + DMR $w/ \, IMFE$   &\textbf{71.68} & \textbf{63.8} &\textbf{71.52} &\textbf{63.97}\\
\bottomrule[0.7pt]
\end{tabular}
}
\label{imfe2}
\end{table}

\textbf{Ablation on IGIM.}
As shown in Fig. \ref{Ablation Study on Time-Frequency mining module}, we have designed several different sub-module compositions for the Intra-modal mining module to achieve modal balance and fusion. The experimental results are presented in Table \ref{ablation on intra-modal}, and the experiments demonstrate that the Time-Frequency Mining sub-module enhances information for sensor modality more effectively than the Spatial Attention sub-module, particularly in the case of modal imbalance (scenario a in Fig. \ref{Ablation Study on Time-Frequency mining module}). 

\begin{figure}[t]
    \centering
    \includegraphics[width=0.8\linewidth]{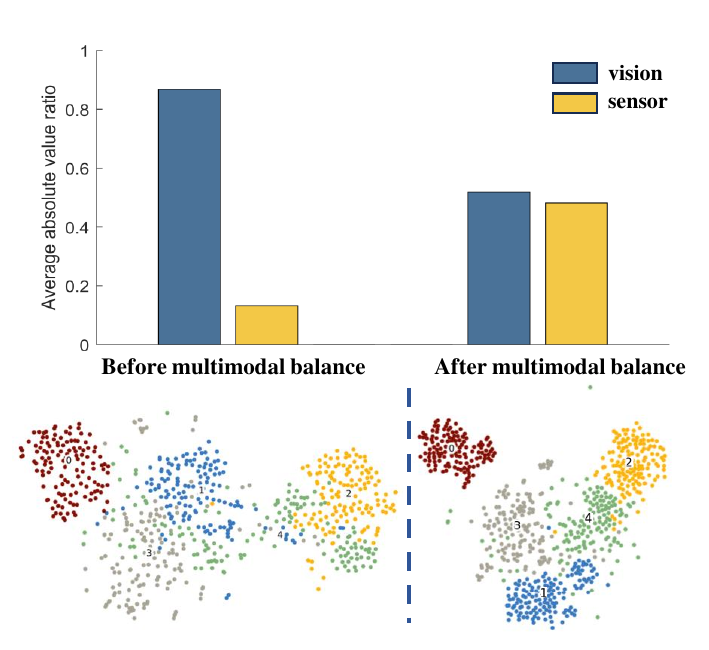}
    \vspace{0em}
    \caption{Effectiveness of proposed IGIM method.The histogram plots the feature map response of the visual and sensor modalities, and the t-SNE plot shows the distribution in the second incremental stage (increment 1 class). Both can be divided into left and right sides for observation. The histogram and t-SNE on the left are before using our multi-modal balancing method, while the right side is after using our method (the color effect is the best).}
    \label{balance}
\end{figure}

We calculate the L1 norm of the visual and sensor feature maps before the CA module in (a) and (c) of the Fig. \ref{Ablation Study on Time-Frequency mining module}, and display the proportions of different modalities in Fig. \ref{balance} (a). From the information in the figure, it can be seen that the representation of sensor-dominant modality features is significantly enhanced, which also verifies that the imbalance in modalities is indeed reduced.
Additionally, we perform t-SNE \cite{van2008visualizing} visualization on sensor modality samples in the incremental process (with an incremental class count of 1). In Fig. \ref{balance} (b), the left image shows the result without using Time-Frequency mining module, while the right image utilizes Time-Frequency mining module. It can be observed from Fig. \ref{balance} (b) that after applying the Time-Frequency mining module, sensor features become more clustered, classification boundaries become clearer, and more useful information is incorporated. The features exhibit significantly increased discriminability.

Additionally, we also conducted ablation experiments on the time-guided method. From the table \ref{ablation on intra-modal}, it can be seen that incorporating prior information on the visual temporal dimension can effectively improve the generalization ability of the multi-modal model.

\begin{table}[t]
\caption{Ablation experiments on the sub-module TFA within the context of IGIM without other methods (Fig. \ref{Ablation Study on Time-Frequency mining module}).}
\vspace{0.5em}
\centering
\renewcommand\arraystretch{1.3}
\resizebox{0.9\linewidth}{!}{
\begin{tabular}{c cccc}
\toprule[0.85pt]
\multirow{2}{*}{}  
&\multicolumn{2}{c}{$T=8$} 
&\multicolumn{2}{c}{$T=4$} \\ 
\cline{2-5} 
& \multicolumn{1}{c}{$\bar{\mathcal{A}}$} 
& \multicolumn{1}{c}{$\mathcal{A}_{T}$} 
& \multicolumn{1}{c}{$\bar{\mathcal{A}}$} 
& \multicolumn{1}{c}{$\mathcal{A}_{T}$} \\ 
\hline
(a)  &92.63 &87.01  &93.26 &89.44\\ 
(b)  &93.84 & 87.84 &94.51 & 90.81\\
(c) Intra-modal miner (base)    &\textbf{94.07} & \textbf{88.68} &\textbf{94.03} &\textbf{91.11}\\
(c)  $w/\, $time-guided  &\textbf{94.80} &\textbf{89.21} &\textbf{95.7} &\textbf{92.02} \\
\bottomrule[0.7pt]
\end{tabular}
}
\label{ablation on intra-modal}
\end{table}

\section{Conclusion}
Because of the nature of data streams, continual learning has garnered significant attention. This paper focuses on maintaining the classification boundary without confusion by preserving the distribution of old samples, thereby retaining old knowledge. Additionally, we address the confusion phenomenon and mitigate forgetting through the IMFE method. As multimodal continuous data streams become increasingly common, multimodal continual learning has attracted widespread attention. This approach not only needs to address forgetting caused by incremental data but also the issue of multimodal generalization. Therefore, this paper first improves multimodal generalization by addressing the imbalance issue inherent in multimodal data through the IGIM. It then follows the intuitive continual learning paradigm of ``learning first, retaining memory second, and avoiding confusion last''. Exploring better and more effective ways to preserve old knowledge in the feature space is a worthwhile endeavor. For instance, the analysis indicates that the significant performance degradation occurs when different dimensions' correlation information is lost in the covariance matrix, highlighting an interesting problem with multiple potential solutions. Furthermore, future work should focus on integrating multimodal continual learning more organically. For example, incremental alignment or fusion mechanisms for multimodal continual learning.

\bibliographystyle{IEEEtran}
\bibliography{IEEEabrv,Multimodal_Egocentric_Dataset}

\end{document}